\pdfoutput=1
\documentclass[11pt]{article}

\usepackage[]{acl}

\usepackage{times}
\usepackage{latexsym}

\usepackage[T1]{fontenc}
\usepackage[utf8]{inputenc}

\usepackage{microtype}

\usepackage{eucal}
\usepackage{amsthm}
\usepackage{graphicx}
\usepackage{caption}
\usepackage{subcaption}
\usepackage{color}
\usepackage{booktabs}
\usepackage{amsmath, amsthm, amssymb}
\usepackage{multirow}
\usepackage{multicol}
\usepackage{xspace}
\usepackage{colortbl}
\usepackage{enumitem}
\usepackage{wrapfig}
\usepackage{algorithm}
\usepackage{algorithmic}
\usepackage{float}
\usepackage{tabularx} 
\usepackage{tablefootnote}
\usepackage{geometry} 
\definecolor{blue}{cmyk}{0.95,0.0,0.2,0.2}

\newcommand{\ours}{\textsc{PL}a\textsc{D}\xspace}

\definecolor{apricot}{rgb}{0.98, 0.81, 0.69}

\DeclareMathSymbol{\shortminus}{\mathbin}{AMSa}{"39}

\newcommand{\blue}[1]{\textcolor{blue}{\small #1}}

\title{\ours: Preference-based Large Language Model Distillation \\ with Pseudo-Preference Pairs}

\author{
  \begin{tabular}{@{}lll@{}}
    Rongzhi Zhang$^1$\footnotemark[1]\hspace*{2cm} Jiaming Shen$^2$\footnotemark[2] \hspace*{2cm} Tianqi Liu$^2$\footnotemark[2]
  \end{tabular}\\
  \begin{tabular}{@{}llll@{}}
  \bf{Haorui Wang$^1$ \hspace*{0.9cm} Zhen Qin$^2$\footnotemark[2] \hspace*{1cm} Feng Han$^2$\footnotemark[2] \hspace*{1cm} Jialu Liu$^2$}
  \end{tabular}\\
  \begin{tabular}{@{}lll@{}}
  \bf{
    Simon Baumgartner$^2$\footnotemark[2] \hspace*{1.2cm} Michael Bendersky$^2$\footnotemark[2] \hspace*{1cm} Chao Zhang$^1$}
  \end{tabular}\\
  $^1$Georgia Institute of Technology, GA, USA \\
  $^2$Google, NY, USA\\
  \texttt{rongzhi.zhang@gatech.edu, jmshen@google.com}
}

\begin{document}
\maketitle
\footnotetext[1]{Work conducted during an internship at Google.}
\footnotetext[2]{Now in Google DeepMind.}
\begin{abstract}
Large Language Models (LLMs) have exhibited impressive capabilities in various tasks, yet their vast parameter sizes restrict their applicability in resource-constrained settings. 
Knowledge distillation (KD) offers a viable solution by transferring expertise from large teacher models to compact student models. 
% Traditional methods like KL divergence can falter due to the student model's restricted expressivity, and relying solely on single teacher outputs may result in poorly calibrated student models. 
However, traditional KD techniques face specific challenges when applied to LLMs, including restricted access to LLM outputs, significant teacher-student capacity gaps, and the inherited mis-calibration issue.
% all affecting student model's learning efficacy.
% While traditional distillation methods can partially help transfer knowledge from large LLM teachers to compact students, these conventional methods face many unique 
In this work, we present \ours, a novel preference-based LLM distillation framework.
\ours exploits the teacher-student capacity discrepancy to generate \emph{pseudo-preference pairs} where teacher outputs are preferred over student outputs.
Then, \ours leverages a ranking loss to re-calibrate student's estimation of sequence likelihood, which steers the student's focus towards understanding the relative quality of outputs instead of simply imitating the teacher.
% propose a novel LLM distillation approach that leverages teacher-student preference pairs, steering the student's focus towards understanding the relative quality of outputs instead of merely replicating teacher outputs.
% for LLM’s distillation that utilizes teacher-student preference pairs, redirecting the student's learning focus towards the relative importance of outputs rather than exact teacher forcing. 
% This methodology provides two key advantages: 
\ours bypasses the need for access to teacher LLM's internal states, tackles the student's expressivity limitations, and mitigates the student mis-calibration issue.
% This method offers dual benefits: it addresses the limitations of student model expressivity and improves sequence ranking calibration, thereby facilitating a more efficient knowledge transfer from teacher to student models. 
Through extensive experiments on two sequence generation tasks and with various LLMs, we demonstrate the effectiveness of our \ours framework. 
\end{abstract}

\section{Introduction}

% \jmshen{Still editing the introduction section below.}
Large language models (LLMs) have shown remarkable abilities across a wide range of tasks~\cite{chatgpt, anil2023palm}.
However, their huge parameter sizes and computational requirements pose significant challenges for practical deployment, especially in environments with limited resources. 
% the extensive parameter counts and computational demands present significant challenges for 
Knowledge distillation (KD) has emerged as a technique for addressing these challenges by transferring insights from a large, sophisticated teacher model to a compact student model with reduced memory footprints and inference costs.
The seminal work~\citep{hinton2015distilling} proposes to train a student model to match the output class distribution of the teacher model.
\citet{kim2016sequence} further extends this idea to the sequence level and teaches the student to directly produce teachers' decoded sequences.
Another line of work~\cite{jiao2019tinybert, wang2020minilm} seeks to align the student model's intermediate-layer representations with the teacher's.
All these approaches employ a teacher-forcing strategy, training the student to fully match the outputs or representations of the teacher model.

% . Typically, the student has fewer parameters
% than the teacher, while maintaining a comparable task-specific performance at a reduced memory footprint and inference cost. 

Applying conventional KD methods to LLMs presents several significant challenges.
First, those LLM teachers are typically only available through API calls.
The absence of direct access to the full output logits or internal states of LLM teachers hinders the implementation of traditional distillation techniques.
Second, the capacity gap between the student model and LLM teachers becomes significantly larger compared to the previous instances when a relatively smaller teacher model was employed.
This disparity exacerbates the student model's limited ability to fully match the teacher LLM's output distribution.
% replicate the complex teacher LLM behaviors through teacher-forcing learning.
Third, as LLMs increase in size, they often encounter a mis-calibration issue~\cite{slic} where sequences that are highly likely according to the model don't necessarily exhibit high quality for target tasks.
Consequently, when a student model is trained to mimic these outputs from the teacher LLM, it inherits this mis-calibration, leading to sub-optimal performance.
Although some recent studies have enhanced the standard teacher-forcing KD paradigm with improved loss functions~\cite{zhang2023not, wen2023f, gu2023knowledge} or learning strategies~\cite{agarwal2023gkd}, these advancements have not yet fully addressed the above challenges, leaving efficient and effective LLM distillation as an open research question.

In this work, we present \textbf{P}reference-based Large \textbf{La}nguage Model \textbf{D}istillation (\ours), a novel framework for distilling LLM knowledge with preference data. \ours is developed based on the following observation: sequences decoded by the teacher model typically surpass the output sequences of the student in quality.
% (c.f. Figure 1).
By sampling outputs from both the teacher and student, \ours generates \emph{pseudo-preference pairs} and calculates a ranking loss that re-calibrates sequence likelihood on the student side.
% Note that we do not require additional human annotations for generating these preference pairs 
This innovation acknowledges the complex teacher-student interaction dynamics within the realm of LLMs and thus shifts the student's learning focus towards understanding the relative quality of different outputs.
% Traditional direct probability matching methods fall short of capturing the nuanced landscape of text generation. 
% We essentially 
Without strictly adhering to teacher forcing, we address the student's inherent limitations in expressivity.

% In response to these challenges, we propose a novel approach to knowledge distillation that leverages teacher-student ranking pairs to compute a calibration loss specifically tailored for LLM generation tasks. 
% Specifically, we sample from both teacher and student to construct pseudo-preference pairs, and calculate a ranking loss to calibrate the sequence likelihood on the student side. 
% Our method recognizes the intricate dynamics of teacher-student interactions in the context of LLMs, where direct probability matching may not fully capture the complexity of the text generation space. 
% By redirecting the student's learning focus towards the relative importance of outputs, rather than strictly adhering to teacher forcing, our method compensates for the student's expressivity limitations. 

% Furthermore, our introduction of a calibration loss bridges the generation quality with its calibrated likelihood, thus explicitly optimizing the generation quality through calibration. 
% Our method not only effectively circumvents the need for internal access to the teacher model, but also introduces a metric-free approach for constructing preference pairs. 
% It naturally exploits the generation quality gap resulting from the model capacity difference and is versatile enough to extend to scenarios where a reward model or ranking metrics are available.

Moreover, the introduction of calibration loss directly ties the quality of generation to its likelihood, allowing for a targeted optimization of output quality through calibration.
This strategy also skillfully bypasses the requirement for teacher model internal access and presents an annotation-free method to construct preference pairs based on the inherent capacity gap between teacher and student models.
\ours is also flexible enough to be applied in scenarios where additional reward models or ranking metrics are available.
This versatility makes it a powerful framework for LLM distillation.

We evaluate \ours on Anthropic helpful dialogue generation ~\cite{bai2022training} and Reddit TL;DR summarization~\cite{stiennon2020learning}.
% and Alpaca-GPT4 instruction following~\cite{peng2023instruction} 
with two different model families LLaMA-2~\cite{Touvron2023Llama2O} and GPT-Neo~\cite{Black2021GPTNeoLS}.
The student model learned by our \ours framework can outperform the one learned with other state-of-the-art KD methods with respect to the win rate of model generations compared to the target sequences. 
We also show that \ours are universally applicable across model families: from PaLM2-L~\cite{anil2023palm} to T5-Large~\cite{2020t5}.

% We use the LLaMA-2-7B as the teacher model, and the GPT-Neo-1.3B~\cite{Black2021GPTNeoLS} as the student model. For evaluation, we report the win rate of the model generation compared to the target sequences as a generic measurement and task-specific metrics. Experiments on three datasets demonstrate distilling with preference can significantly improve the student model performance.

\textbf{Contributions.} 
The major contributions of this work are summarized as follows: 
% In summary, we make the following contributions:
(1) We propose \ours, a novel framework that distills LLMs with preference data; 
% Proposed a new preference distillation paradigm for generation tasks of LLMs;
(2) We present a metric-free approach to construct pseudo-preference pairs without human annotations;
(3) We facilitate the LLM distillation with an explicit calibration objective and improve the student model's generation capability;
(4) We conduct comprehensive experiments on multiple tasks with different-sized teacher models to demonstrate the effectiveness of \ours.
\section{Related Work}

\noindent \textbf{Sequence Knowledge Distillation.}
Knowledge distillation (KD) is first proposed in~\citep{bucilua2006model} to compress the large models to smaller, faster models without a significant performance drop. 
\citet{hinton2015distilling} generalizes this technique by introducing a temperature parameter to smooth the teacher model prediction.
SeqKD~\citep{kim2016sequence}, initially targeting the neural machine translation task, extends the scope of KD from multi-class classification to sequence generation and learns a distill student model to generate a sequence holistically.
% Initially introduced for model compression \citep{bucilua2006model}, KD was generalized with the introduction of temperature scaling to soften the teacher's outputs~\citep{hinton2015distilling}. 
% \citet{kim2016sequence} specifically applied KD to sequence generation tasks, improving the training of student models in neural machine translation.
Further developments have seen the incorporation of contrastive learning \citep{tian2019contrastive} and patient distillation techniques \citep{sun2019patient}, where the student learns from multiple layers of the teacher model. Transformer-specific distillation methods have also been proposed \citep{sanh2019distilbert, jiao2019tinybert}, focusing on attention and hidden state alignment for efficient knowledge transfer. These advancements underscore the ongoing efforts to refine SeqKD for natural language processing tasks, balancing model size with linguistic performance.

\smallskip
\noindent \textbf{Learning from Preference Data.}
The pivotal Reinforcement Learning from Human Feedback (RLHF) framework~\citep{rlhf} first uses preference data to fit a reward model and then fine-tunes the LM to maximize the given reward using reinforcement learning algorithms. 
In practice, however, using RL to directly fine-tune LMs are challenging, incurring significant computational costs and requiring extensive hyperparameter tuning. 
To mitigate this issue, DPO~\citep{dpo} proposes to directly train the policy LM using the pairwise logistic loss without explicitly fitting a reward model.
SLiC~\citep{slic} and RRHF~\citep{rrhf} adopt the pairwise hinge loss to train the policy LM.
RSO~\citep{rso} further introduces the statistical rejection sampling method to improve DPO by addressing the distribution drift issue.
More recently, LiPO~\cite{Liu2024LiPOLP} leverages a ranked list of responses (either rated by humans for ranked by models~\cite{Qin2023LargeLM}) for LLM alignment.
All these studies require either human annotations or external reward models to obtain the preference data and focus on aligning a single model with human preference.
Additionally, some works have attempted to explain the distillation mechanisms from multiple perspectives\cite{lopez2015unifying,lopes2017data,zhang2020seqmix,menon2021statistical,zhang2022prboost}, but none have yet bridged the preference learning and knowledge distillation, while it is a natural approach given the capacity gap between teacher and student models.
In this work, we focus on distilling a teacher model into a student model with self-supervised preference pairs.

\smallskip
\noindent \textbf{LLM Distillation.}
Recent efforts in the distillation of Large Language Models (LLMs) have introduced innovative approaches to refine the knowledge transfer process. 
\cite{liang2023less} employ a task-aware, layer-wise distillation method which effectively tackles the challenges of tasking student models mimicking the hidden representation of a much larger teacher, but it requires access to the teacher model's intermediate layers, whereas our research focuses on the scenarios where only the teacher model's final sequence-level output is available, which is common in the context of LLMs.
\cite{zhang2023lifting} seek to bridge the capacity gap between student models and teacher models by deploying the mixture of experts (MoE) architecture on the student side, thereby increasing its capacity. We alternatively address the capacity gap by shifting from traditional teacher-forcing distillation to a preference-based distillation, leveraging synthetic preference data and avoid memory footprint overhead increased by complex architectures of the student model. 

Another line of works leverages additional knowledge to improve LLM distillation. For instance, ``Distillation step-by-step'' by~\citep{Hsieh2023DistillingSO} enriches the student model training by integrating LLM output rationales, aiming for a deeper understanding and replication of the teacher model's reasoning pathways. 
Similarly, \citet{Fu2023SpecializingSL} advance the methodology by focusing on LLM output Chains of Thought (CoT), engaging in per token distribution matching to capture nuanced decision-making processes. 
\citet{Shridhar2023DistillingMR}~ take a specialized approach towards reasoning tasks, proposing a dual-student model framework where one student model is dedicated to problem decomposition and the other to solving the identified subproblems, facilitating a more segmented yet comprehensive distillation strategy.
Additionally, \citet{zhang2023not} also introduce a non-teacher forcing distillation approach, where the teacher output is perturbed to get a proxy teacher with a distribution closer to the ground truth.
Despite these advancements, none of these works incorporate preference data into the distillation process, highlighting a key contribution of this study.

\section{Preliminaries}
\subsection{Auto-Regressive Text Generation}
We denote the input and output sequence as \( x \) and \( y \), respectively. Let \( V \) denote the vocabulary comprising of \( M \) tokens, \( y_{<n+1} = (y_1, y_2, \ldots, y_n) \) denote the generated output sequence up to the \( n \)-th token, and \( L_y \) denote the length of sequence \( y \). An auto-regressive policy \( p(.|y_{<n}, x) \) outputs a next-token probability distribution over all tokens in \( V \), conditioned on the input \( x \) and output sequence \( y_{<n} \). 
The probability  \( p(y_n | x) \) of predicting the \( n \)-th token in sequence $y$ is determined by a softmax function with temperature \( \gamma \) as follows:
\begin{equation}
\small
    p(y_n|x) = \frac{\exp(z_{n}/\gamma)}{\sum_{i=1}^{M} \exp(z_i/\gamma)},
\end{equation}
where \( z_i \) is the logit score for the $i$-th token in \( V \). 
Higher values of \( \gamma \) introduce more randomness, while a lower value makes the output more deterministic by favoring the most probable words.
In conditional language generation tasks, the model produces a response $y$
% \( y = \{y_t\}_{t=1}^{T} \) 
conditioning on a prompt \( x \) sampled from the distribution \( p_x \). 
The output sequence \( y \) is sampled from the generative model in an auto-regressive manner as described above.

% \subsection{Knowledge Distillation in Language Modeling}
\subsection{Sequence Knowledge Distillation}
We approach knowledge distillation (KD) as an optimization problem to minimize the difference between a fixed teacher model output distribution \( p(y|x) \) and a student model output distribution \( q_\theta(y|x) \), parameterized by \( \theta \). 
The standard KD method for generative models aims to minimize the forward KL divergence:
\begin{equation}
\small
    KL[p||q] = \mathbb{E}_{x\sim p_x, y\sim p'} \log \frac{p(y|x)}{q_\theta(y|x)},
\end{equation}
where \( p' \) represents either the real data distribution for word-level KD or the teacher distribution \( p \) for sequence-level KD. 
However, \( KL[p||q] \) tends to overestimate the low-value regions of \( p \) in language generation tasks when \( q_\theta \) lacks the expressiveness to cover all the modes of \( p' \). This issue is particularly pertinent for LLMs that perform a variety of tasks in a generative manner, since the low-capacity student models are unable to imitate the complex language generation distribution of their teacher models or that of humans perfectly.

% \section{Distilling from Preference Data}
\section{\ours: Preference-based Large Language Model Distillation}

\begin{figure*}
    \centering
    \includegraphics[width=\textwidth]{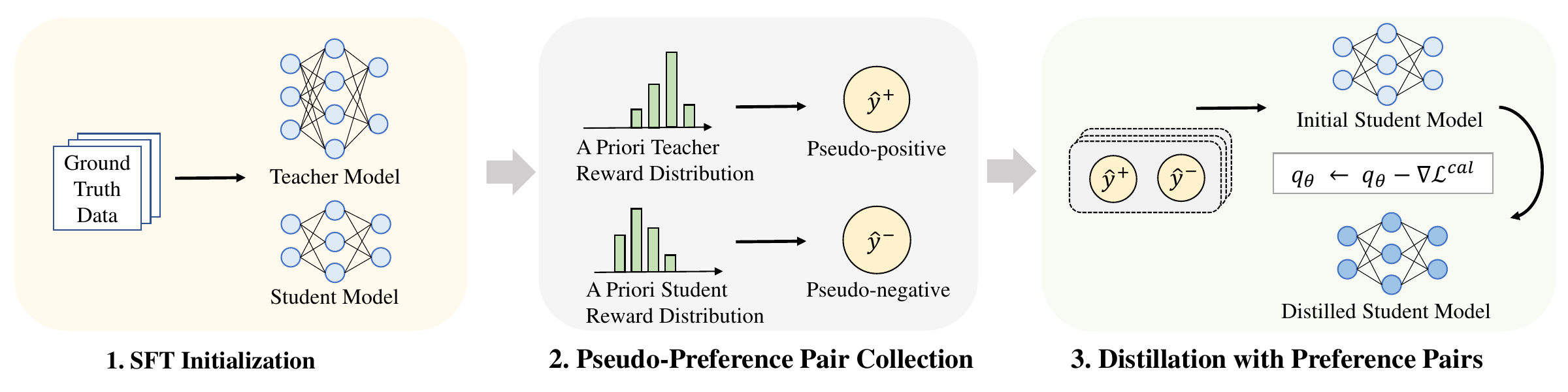}
    \caption{Our \ours framework starts with separate SFT processes for both the teacher and student models. The best checkpoint of the teacher model is selected based on the win rate over targets, while the student model went through the whole SFT for initialization. In the next stage, we generate pseudo-preference data by sampling generation pairs from the teacher and the student. The student model then undergoes preference distillation using this pseudo-preference data to produce a distilled student model.}
    \label{fig:enter-label}
\end{figure*}

\subsection{Framework Overview}
Our proposed \ours framework starts from the supervised fine-tuning (SFT) phase for both the teacher and the student models. 
First, the teacher model undergoes SFT to optimize its parameters for the target tasks. 
The student model is similarly fine-tuned to prepare for the subsequent distillation phase. 
% To calibrate the likelihood of the student generation in the following distillation phase, we first construct pseudo-preference pairs. 
Then, we construct pseudo-preference pairs to calibrate the likelihood of the student generation.
Specifically, we conduct inference on a distillation set, which consists of task-specific inputs without corresponding target outputs. 
The sampled generations from both the teacher and student models are used to form the pseudo-preference data, where we assume the teacher output is preferred over student output due to their capacity difference. 
Finally, with the pseudo-preference data, we implement a calibration loss to optimize the generation quality in the distillation phase explicitly. The final distilled student model is evaluated by win rate and ROUGE scores to compare with its SFT predecessor and other baselines. 
We outline the proposed \ours framework in Figure~\ref{fig:enter-label}.

\subsection{Pseudo Preference Pairs Generation}
% The concept of pseudo-preference data is central to our \ours framework, offering a practical and efficient alternative to using ground truth preference pairs. Ground truth preferences typically rely on human annotations or state-of-the-art model inferences, which can be prohibitively costly and time-intensive to obtain. Our approach circumvents this by leveraging the inherent quality disparity between the teacher and student models — a byproduct of the natural capacity gap that exists between them. This gap justifies the assumption that the teacher's generation is of superior quality when compared to that of the student.

Our \ours framework pivots around the concept of pseudo-preference data,
% is central to our \ours framework, 
offering a practical and efficient alternative to human annotated preference pairs. 
Traditional approaches to preference learning, such as Deep Preference Optimization (DPO)~\cite{dpo} and Reinforcement Learning from Human Feedback (RLHF)~\cite{rlhf}, rely on preference data obtained through costly human annotations or inferences from state-of-the-art models. These methods, while effective, are prohibitively expensive and time-consuming for large-scale applications. 

To mitigate these issues, we capitalize on the reliable assumption that the teacher model's generative quality supersedes that of the student due to its greater capacity. 
Consequently, we can generate pseudo-preference pairs by sampling outputs from both models on the distillation set and assuming the teacher output is preferred over the student output.
% teacher-student rank --  The capacity gap between the teacher and the student justifies the assumption of the quality difference of their generation. 
Formally, the generation process for a given input \( x \) from the distillation set can be expressed as:
\begin{equation}
    (\hat{y}_{+}, \hat{y}_{-}) \mathrel{\mathop:}= (\hat{y}^T, \hat{y}^S) = (p(y|x), q_{\theta}(y|x))
\end{equation}
where \( \hat{y}^T \) and \( \hat{y}^S \) represent the generations from the teacher and student models, respectively. We then construct the preference pairs \( (\hat{y}_{+}, \hat{y}_{-}) \), with \(\hat{y}_{+} \mathrel{\mathop:}= \hat{y}^T \) signifying the teacher's higher-quality output and \(\hat{y}_{-} \mathrel{\mathop:}= \hat{y}^S \) indicating the student's output. These pseudo-preference pairs provide a cost-effective and scalable alternative to human-annotated or AI-annotated data, enabling the student model to redirect the teacher to the relative importance of learning.
% compensating for the limited expressivity of the student.

\subsection{Distillation with Preference Pairs}
The distillation with preference pairs is formalized through a calibration loss, designed to bridge the student's generative quality to its likelihood. We employ two types of losses: the ranking calibration loss, $L^{\text{cal}}_{\text{rank}}$, and the margin calibration loss, $L^{\text{cal}}_{\text{margin}}$.

The ranking calibration loss is defined as:
\begin{equation}
\small
    L^{\text{cal}}_{\text{rank}} = \max(0, \beta - \log P_\theta (\hat{y}_+ | x) + \log P_\theta (\hat{y}_- | x)),
\label{eq:cal_rank}
\end{equation}
where $\beta$ is a margin hyper-parameter, and $\hat{y}_+$ ($\hat{y}_-$) represents the teacher (student) outputs from the pseudo preference pairs, respectively. 
This loss encourages the student model to increase the probability of the preferred output while decreasing the likelihood of the less preferred one.

The margin calibration loss is introduced to refine the student's output by considering a scoring function $s$, which provides an additional quality measure for the generated sequences:
\begin{equation}
\small
\begin{split}
    L^{\text{cal}}_{\text{margin}} & = \max(0, \beta (s(y;\hat{y}_+; x) - s(y;\hat{y}_-;x)) \\
    & - \log P_\theta (\hat{y}_+ | x) + \log P_\theta (\hat{y}_- | x))
\end{split}
\label{eq:cal_margin}
\end{equation}
In this equation, $s(y;\hat{y}_+; x)$ and $s(y;\hat{y}_-; y;x)$ represent the scores of the preferred and less-preferred outputs, respectively. This loss function penalizes the student model when the score of the less-preferred output is too close to that of the preferred one, promoting a distinction between high and low-quality generations. In practice, we choose the learning objectives based on the performance on the validation set.

By leveraging the calibration loss in conjunction with pseudo-preference pairs, our method enables an effective distillation process, fostering a student model that not only performs well on text generation tasks but also exhibits calibrated confidence in its outputs.
We summarize the framework in Algorithm.\ref{alg:pkd}.

\begin{algorithm}[!t]
\small
\caption{Teacher-Student Knowledge Distillation with Calibration Loss}
\begin{algorithmic}[1]
\STATE \textbf{Require:} Teacher model $p$, student model $q_{\theta}$, SFT dataset $\mathcal{D}_0$ with labeled target sequences, and distillation set $\mathcal{D}$.
\STATE \blue{// Step 1: \emph{Initialization}}
\STATE Learn SFT teacher $p$ and initial student $q_{\theta}$ on $\mathcal{D}_0$.

\STATE \blue{// Step 2: \emph{Pseudo-Preference Pair Construction}}
\FOR{each input sequence $x \in D$}
    \STATE Sample teacher output $y_T = p(y|x)$.
    \STATE Sample student output $y_S = q_{\theta}(y|x)$.
    \STATE Make pseudo-preference pairs $(\hat{y}_{+}, \hat{y}_{-}) \mathrel{\mathop:}= (y_T, y_S)$.
\ENDFOR
\STATE \blue{// Step 3: \emph{Student Distillation with Preference Pairs}}
\FOR{each batch in $\mathcal{D}$}
    \STATE Compute loss $\mathcal{L}^{cal}$ via Eq.~\ref{eq:cal_rank} or \ref{eq:cal_margin}.
    \STATE Update student model $q_{\theta} \leftarrow q_{\theta} - \nabla \mathcal{L}^{cal}$.
\ENDFOR
\STATE \textbf{Output:} Distilled student model $q_{\theta}$.
\end{algorithmic}\label{alg:pkd}
\end{algorithm}

\section{Experiments}
\subsection{Experiment Setup}
\noindent\textbf{Datasets.} 
We conduct experiments on the following datasets: 
(1) \textbf{TL;DR}~\cite{stiennon2020learning} which comprises around 140k Reddit posts along with their TL;DR summarizations, providing a rich source for training and evaluating text summarization models, and (2) \textbf{Anthropic-HH}~\citep{bai2022training} which initially designed for training preference models in a dialog system with Reinforcement Learning from Human Feedback (RLHF). We use its helpful slice for experiments.
More detailed dataset statistics are listed in Appendix~\ref{app:dataset_stats}.
% It is not intended for the supervised training of dialogue agents, as such training may result in harmful models.\\
% \begin{itemize}
%     \item \textbf{HH-RLHF} \citep{bai2022training}: This dataset is designed for training preference models (or reward models) for Reinforcement Learning from Human Feedback (RLHF). It is not intended for the supervised training of dialogue agents, as such training may result in harmful models.
    
%     \item \textbf{TL;DR} \cite{stiennon2020learning}: The Reddit TL;DR dataset comprises Reddit posts along with their summarizations, providing a rich source for training and evaluating text summarization models.
    
%     % \item \textbf{Alpaca-GPT4}: This dataset includes English instruction-following data, generated by GPT-4 using Alpaca prompts, and is used for fine-tuning Large Language Models (LLMs).
% \end{itemize}

\smallskip
\noindent\textbf{Models.}
We evaluate two model families in our main experiments:
(1) \textbf{LLaMA-2 Models}~\cite{Touvron2023Llama2O} include LLaMA-2-13B as the teacher model and LLaMA-2-7B as the student model, and
(2) \textbf{GPT-Neo Models}~\cite{Black2021GPTNeoLS} include GPT-Neo-2.7B as the teacher model and GPT-Neo-1.3B as the student model.
Besides, we also extend \ours to PaLM-2~\cite{anil2023palm} and T5 models~\cite{2020t5} to show its broad applicability.

\smallskip
\noindent\textbf{Baseline Methods.} 
We compare \ours~ to both classic KD techniques and LLM KD techniques, including
(1) \textbf{Standard KD}~\citep{hinton2015distilling}: The foundational knowledge distillation technique that trains a student model to replicate the teacher model's output distributions;
(2) \textbf{SeqKD}~\citep{kim2016sequence}: An extension of standard KD to sequence generation that distills the student model directly on the teacher's generations;
(3) \textbf{f-distill}~\citep{wen2023f}: A framework that addresses KL divergence's the mode averaging and collapsing problems by minimizing a symmetric f-divergence; and
(4) \textbf{MiniLLM}~\citep{gu2023knowledge}: A framework that distills LLMs into their smaller counterparts using reverse KL divergence.

    % The implementation is available at \url{https://github.com/microsoft/LMOps/tree/main/minillm}.

    % \item \textbf{GKD} \citep{agarwal2023gkd}: Presents an on-policy imitation method for incorporating teacher feedback, compared with ImitKD, f-distill, SeqKD, and Standard KD (token-level KD).
    
    % \item \textbf{DPO} \citep{rafailov2023direct}: Direct Preference Optimization for training models in alignment with human preferences.
\smallskip
\noindent\textbf{Evaluation Schemes.}
We calculate the win rate as the major metric for model evaluation. Win rate is defined as the ratio of preferred generation compared to the target text. Specifically, we deploy a task-specific reward model and use the human-written reference sequence as the target. Both the reward model and the reference sequence are pre-provided in the open-source community and the datasets. Besides, we also provide ROUGE scores for reference.
More details are provided in Appendix~\ref{app:model}.

% \smallskip
% \noindent \textbf{Evaluation Model}: We take a DeBERTa model \footnote{https://huggingface.co/OpenAssistant/reward-model-deberta-v3-large-v2} trained on related tasks as the reward model. We detail the above models in Appendix~\ref{app:model}.\\

\smallskip
\noindent\textbf{Implementation Details.}
We split each task's original training set into two equal shards for training and distillation. 
Given the training set $\mathcal{D}_0$, we do 1-epoch fully supervised training to get the teacher model and initialize the student model. 
We run inference and construct teacher-student pseudo-preference pairs for the distillation set $\mathcal{D}$. The original test set is kept for evaluation. For model training, the learning rate is  $1e^{-4}$ with a linear scheduler, and the per-device batch size is 8. We also use LoRA \citep{hu2021lora} for all experiments for training efficiency. Specifically, we set LoRA rank as 8, LoRA dropout as 0.1, and LoRA alpha as 32. We list more details in Appendix~\ref{app:hp}.

% \subsection{Main Results}
% \input{table/main_res}

% \subsection{Analysis}
% \noindent\textbf{Ablation study 1 - real reward pair?}

% \noindent\textbf{Ablation study 2 - general-purpose model for win rate calculation}

% \noindent\textbf{Ablation study 3 - hpara study}

% \noindent\textbf{Other metrics such as Rouge/GLUE/normalized rewards}

% \noindent\textbf{Length Normalization}

% \noindent\textbf{Scaling w/ model or data (number of sampling)}
% \begin{figure*}[!htb]
%     \centering
%     \includegraphics[width=\textwidth]{figure/scaling.pdf}
%     \caption{Scaling with the number of parameters of student models.}
%     \label{fig:enter-label}
% \end{figure*}

% \noindent\textbf{Compared to learning from preference, i.e., DPO}

% \noindent\textbf{Case Study}
% \begin{figure*}
%     \centering
%     \includegraphics[width=\textwidth]{figure/case_study.pdf}
%     \vspace{-30pt}
%     \caption{The case study on the TL;DR dataset. In the original text, critical details are highlighted in green. Similarly, in the student's summary, critical details are also highlighted in green, with redundant parts in yellow and factual errors in red. Compared to the summaries produced by the SFT student model, those generated by the student model distilled with \ours are notably more concise and accurate, capturing a broader range of key points.}
%     \label{fig:enter-label}
% \end{figure*}

%----------------------Skeleton Below--------------------%
\subsection{Main Results}
\definecolor{LightBlue}{HTML}{DEEBF7}

\begin{table*}[!htb]
\centering
\resizebox{15cm}{!}{
\begin{tabular}{c|c|ccccc|ccccc}
\toprule
\multirow{2}{*}[-0.2em]{\textbf{Model}} 
& \multirow{2}{*}[-0.2em]{\textbf{Method}} 
& \multicolumn{5}{c|}{\textbf{TL-DR} \quad  | T-S WR = 64.86} 
& \multicolumn{5}{c}{\textbf{Anthropic-HH}  | T-S WR = 53.97} \\
\addlinespace[0.2em] \cline{3-12} \addlinespace[0.2em]
& & \textbf{Word Count} & \textbf{R-1} &\textbf{R-2} &\textbf{R-L} & \textbf{RM WR (\%)} & \textbf{Word Count} & \textbf{R-1} &\textbf{R-2} &\textbf{R-L} & \textbf{RM WR (\%)}\\
\midrule
\multirow{8}{*}[-0.2em]{\textbf{LLaMA-2}} & Teacher & 25.51	& 33.14 & 11.40 & 25.31 & 45.28 & 67.78 &  25.15 & 6.50 & 16.32 & 26.95 \\
& Initial Student & 27.24&	26.68&	8.98&	20.60&	33.42 & 67.80 &  24.96 & 6.48 & 16.26 & 25.63 \\
& SFT & 25.77&	30.06&	9.95&	22.97&	35.13 & 67.64 &  25.33 & 6.66 & 16.51 & 25.52\\
& Standard KD & 26.32 & 31.45 & 10.04 & 23.78 & 37.70 & 67.60 & 25.06 & 6.57 & 16.31 & 25.83\\
& SeqKD & 25.67 & 31.67 & 10.42 & 24.20 & 37.92 & 67.73 & 25.02 & 6.51 & 16.28  & 26.24\\
& f-distill & 25.98 & 32.09 & 10.65 & 24.42 & 38.35 & 68.24 & 25.23 & 6.72 & 16.40 & 25.98\\
& MiniLLM & 25.80 & 31.32 & 10.48 & 24.25 & 38.07 & 68.05 & 25.42 & 6.76 & 16.53 & 26.61 \\
& Ours& \cellcolor{LightBlue}26.02 & \cellcolor{LightBlue}31.46 & \cellcolor{LightBlue}10.79 & \cellcolor{LightBlue}24.91 & \cellcolor{LightBlue}40.46 & \cellcolor{LightBlue}67.93 & \cellcolor{LightBlue}25.54 & \cellcolor{LightBlue}6.82 & \cellcolor{LightBlue}16.70 & \cellcolor{LightBlue}27.74\\
\midrule
\multirow{2}{*}[-0.2em]{\textbf{Model}} 
& \multirow{2}{*}[-0.2em]{\textbf{Method}} &  \multicolumn{5}{c|}{\textbf{TL-DR} \quad  | T-S WR = 60.80} 
& \multicolumn{5}{c}{\textbf{Anthropic-HH}  | T-S WR = 55.23} \\
\addlinespace[0.2em] \cline{3-12} \addlinespace[0.2em]
& & \textbf{Word Count} & \textbf{R1} &\textbf{R2} &\textbf{RL} & \textbf{RM WR (\%)} & \textbf{Word Count} & \textbf{R1} &\textbf{R2} &\textbf{RL} & \textbf{RM WR (\%)}\\
\midrule
\multirow{8}{*}[-0.2em]{\textbf{GPT-Neo}} & Teacher& 25.81&	29.64&	8.82	&22.22	&15.35 & 72.55	&21.47&	4.12&	13.66	&	15.52 \\
& Initial Student & 26.35&	29.04&	8.34&	21.66&	11.51  &73.53	&19.77&	3.36&	12.67&	8.73 \\
& SFT Student &25.88&	29.08&	8.44&	21.76&	11.40& 73.49&	18.51&	2.76&	11.93&	8.79 \\
& Standard KD & 26.50 & 29.15 & 8.51 & 21.84  & 12.47 & 73.09 & 20.35 & 3.56 & 12.78 & 8.94 \\
& SeqKD&   26.14& 29.23 & 8.57 & 21.82 & 13.18 & 73.11 & 20.74 & 3.75 & 13.05 & 9.25\\
& f-distill & 25.73 & 29.34 & 8.62 & 21.91  & 13.55 & 72.68 & 20.67 & 3.84 & 13.74  & 9.66\\
& MiniLLM & 25.86 & 29.37 & 8.62 & 22.03 & 13.73 & 72.38 & 20.90 & 3.80 & 13.56 & 9.81 \\
& Ours& \cellcolor{LightBlue}25.77 & \cellcolor{LightBlue}29.50 & \cellcolor{LightBlue}8.69 & \cellcolor{LightBlue}22.08 &  \cellcolor{LightBlue}14.76 & \cellcolor{LightBlue}72.76 & \cellcolor{LightBlue}21.12 & \cellcolor{LightBlue}3.82 & \cellcolor{LightBlue}12.83 &  \cellcolor{LightBlue}10.38 \\
\bottomrule
\end{tabular}}
\caption{Main results with the LLaMA-2 and GPT-Neo models. For the LLaMA-2 group, the teacher model is LLaMA-2-13B, and the student model is LLaMA-2-7B. For the GPT-Neo group, the teacher model is GPT-Neo-2.7B, and the student model is GPT-Neo-1.3B. WR stands for win rate, and RM refers to a task-specific reward model. The teacher-student win rate (T-S WR) on the distillation set evaluated by a reward model is provided for reference.}
\label{tab:main_res_merge}
\vspace{-8pt}
\end{table*}

Table~\ref{tab:main_res_merge} presents our main experiment results.
\ours exhibits a notable capacity for closing the teacher-student performance gap.
Compared to the initial student and the SFT baseline, the student model learned by \ours~ not only significantly improves the win rate but also enhances the ROUGE scores.
Impressively, for the Anthropic-HH task, the student model learned by \ours even surpasses the teacher model in terms of the win rate (student's 27.74\% win rate against the teacher's 26.96\% win rate). 

In comparison to KD baselines, our approach consistently delivers superior performance across different settings. This is evident from our method exhibiting the highest win rate among all the student models distilled by baseline methods. We also count the generation length by the number of words to verify if the quality improvement comes from the output verbosity. We notice that the word count across different methods is relatively stable. We thus further conduct an experiment in Section~\ref{sec:len_perform} to justify the performance on different generation length ranges. 

Moreover, the Teacher-Student Win Rate (T-S WR) emerges as a critical determinant in the final performance. For the stronger teachers like the LLaMA teacher and the GPT-Neo teacher on the TL;DR task, the distilled student improves by relatively larger margins. Conversely, when learning from a mediocre teacher model, as in the case of the Anthropic-HH task with a lower T-S WR, the distilled student models show only marginal advancements. This highlights the importance of the teacher model's quality in the distillation process and its impact on the student model's ultimate efficacy.

% \subsection{Analysis}
% In our extensive analysis, we delve into the performance nuances of our proposed distillation method, exploring various facets of the model and training process.\\

\subsection{Impact of Real Preference Pairs}
\begin{figure*}
    \centering
    \includegraphics[scale=0.4]{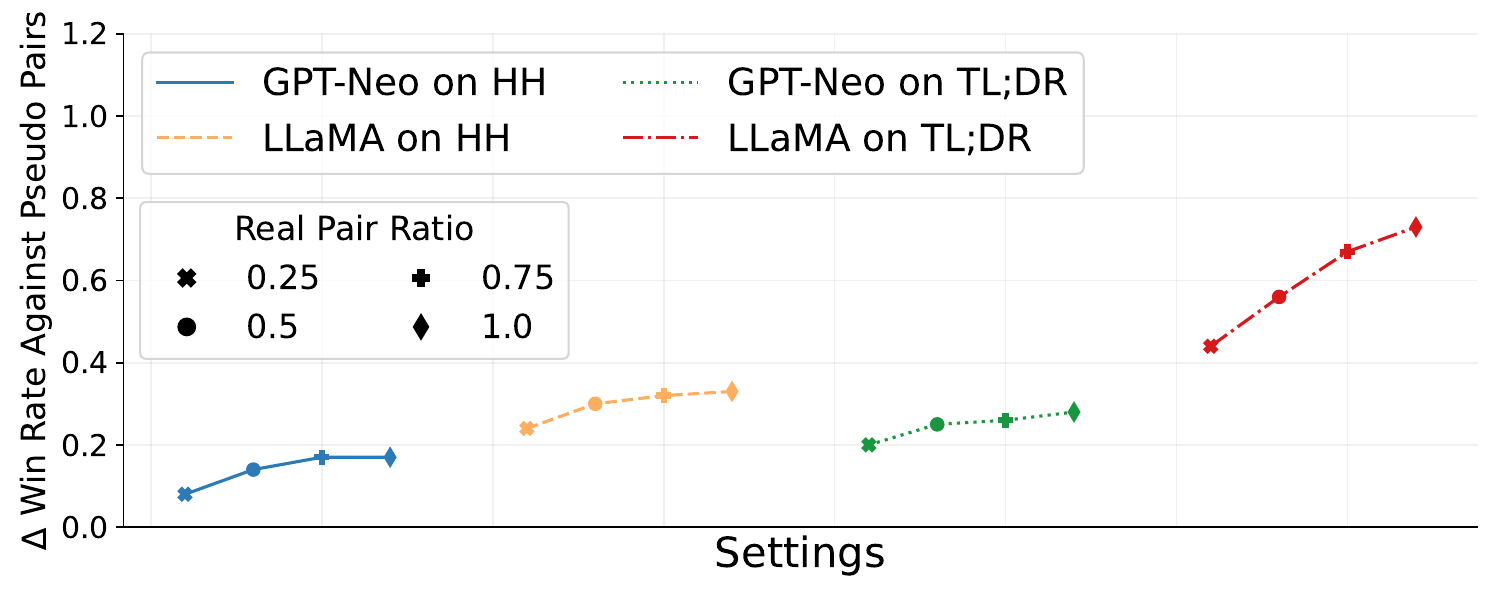}
        \vspace{-10pt}
    \caption{Comparison between using real preference pairs compared to using pseudo preference pairs.}
    \vspace{-10pt}
    \label{fig:ratio}
\end{figure*}
In this set of experiments, we investigate the effect of using real preference pairs compared to pseudo-preference pairs.
We use both the LLaMA-2 model and the GPT-Neo model on TL;DR and Anthropic-HH datasets, each setting comes with varying ratios of real to pseudo preference pairs. Here we use a reward model to evaluate all the pairs and adjust the ranking within each pair based on their reward values.

In Figure~\ref{fig:ratio}, we observe a slight improvement in the win rate against using pseudo pairs as the ratio of real preference pairs increases. This trend is evident across both models and settings, albeit to varying degrees. The win rate improvement is more pronounced in the LLaMA-2 models on the TL;DR dataset, indicating a model-specific benefit from real preference data.

Considering the original improvements from using pseudo-preference pairs, the gain of replacing the pseudo-preference pairs with real ones is marginal.  
While the use of real preference pairs does yield improvements in model performance, it also incurs additional human annotation costs. This presents a trade-off scenario where the gains from real preference data are weighed against the resource expenditure associated with their use.

In conclusion, the experiment outcomes support the reliability of employing pseudo-preference pairs in LLM distillation, supplemented by the proposed calibration objective. Despite the slight edge provided by real preference pairs, the cost-effective and time-efficient nature of pseudo pairs makes them a viable alternative. This study suggests that pseudo-preference pairs, when properly constructed, can serve as a practical proxy to real preference data without significantly compromising the learning efficacy of language models.

\subsection{Performance in Length Ranges}\label{sec:len_perform}
\begin{figure}
    \centering
    \includegraphics[scale=0.28]{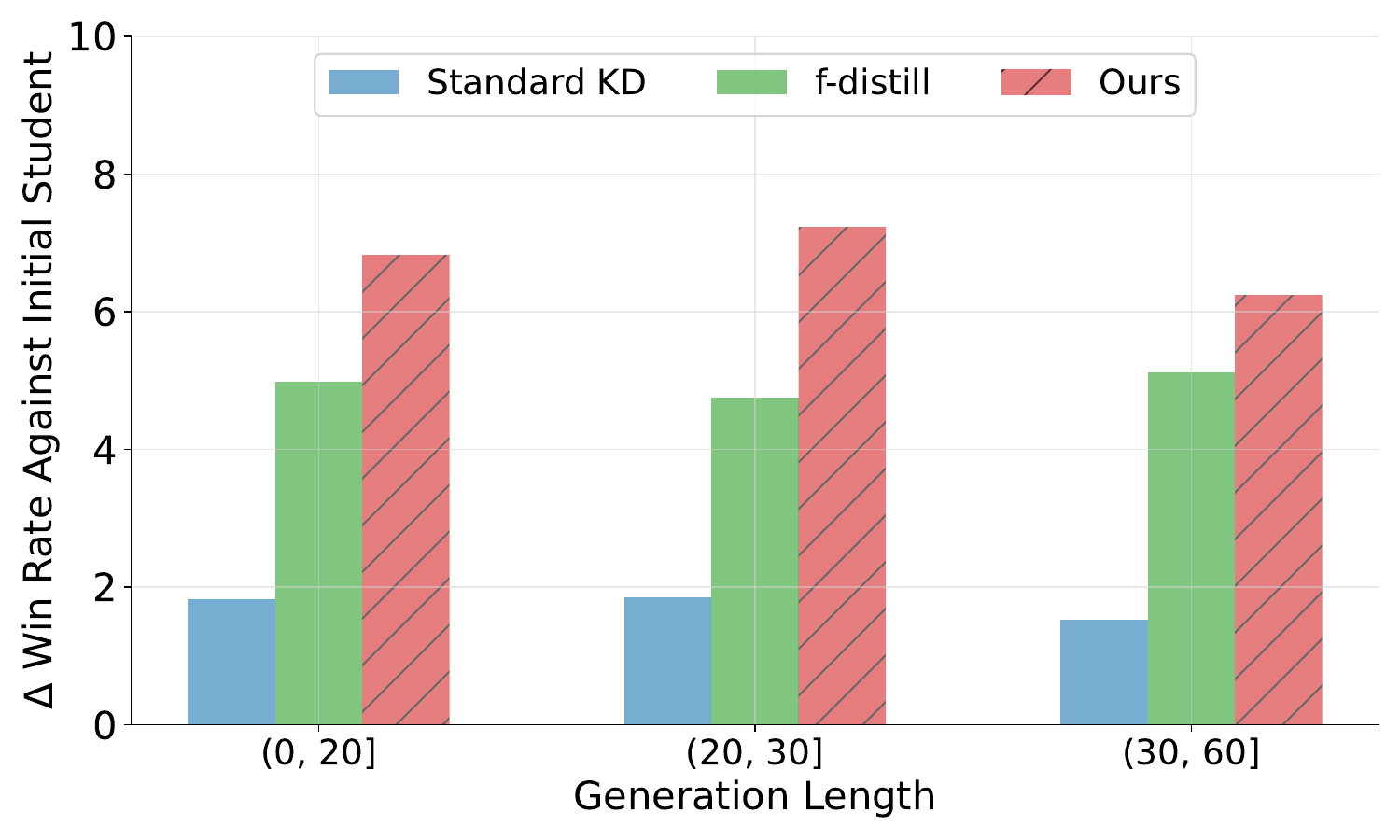}
    \vspace{-5pt}
    \caption{The $\triangle$ win rate against the initial student v.s. the length of student model generation. Experiments are conducted on TL-DR with LLaMA-2.}
        \vspace{-10pt}
    \label{fig:len_wr}
\end{figure}
In this experiment, we examine the correlation between generation length and win rate. This experiment is conducted on the TL-DR task using LLAMA-2 models. Figure.~\ref{fig:len_wr} indicates all methods maintain a relatively stable win rate improvement across varying generation lengths. 
 It also illustrates that our method consistently leads to better performance across all generation lengths of student models, with a notable peak within the (20, 30] interval, achieving the highest win rate improvement to the initial student and the highest margin over the baselines. Considering the average generation length is around 26 on this task, this suggests that our approach is particularly effective in the most commonly encountered scenario within the TL-DR task, and the enhancement at this central interval is especially valuable given its prevalence in the dataset.\\

\vspace{-5pt}
\subsection{Scaling with Distillation Data}
\begin{figure}
    \centering
    \includegraphics[scale=0.25]{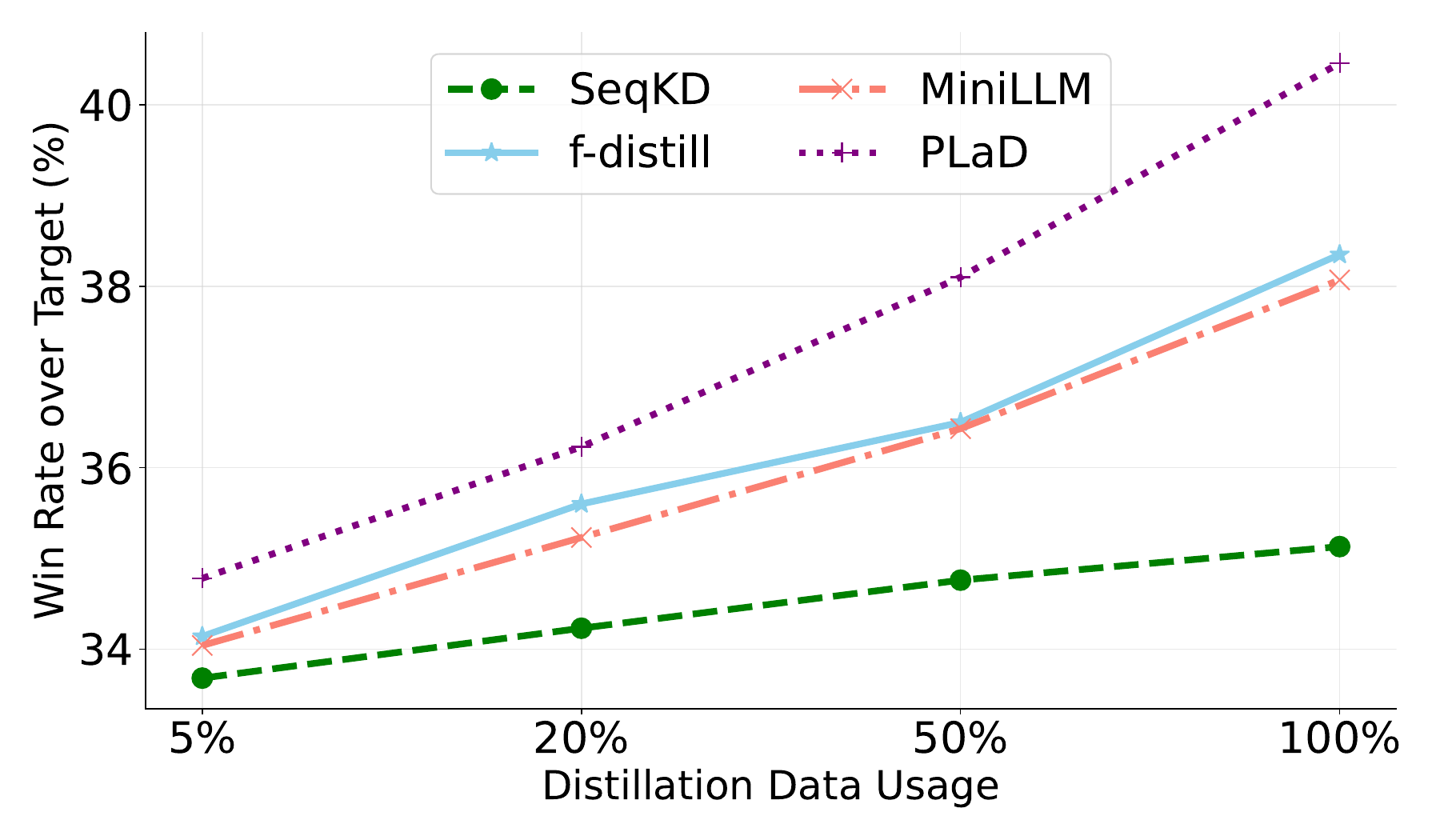}
    \vspace{-22pt}
    \caption{Scaling properties of the distillation data. }
    \label{fig:scaling}
    \vspace{-10pt}
\end{figure}

We present a scaling analysis in Figure~\ref{fig:scaling} to demonstrate how our distillation method's effectiveness varies with changes in the amount of distillation data. We keep the same setting as the LLaMA-TL;DR group in the main experiments, and perform distillation with different distillation data usage. 

The figure illustrates that as the percentage of distillation data increases from 5\% to 100\%, all methods show an upward trend in win rate over the target, indicating improved model performance with access to more distillation data. Notably, \ours~ demonstrates a steep improvement curve, outperforming the other methods, particularly at higher data usage levels. This suggests that \ours~ is highly efficient in refining the student model.

% \noindent\textbf{Comparison with Learning from Preference.}
% We draw comparisons with Direct Preference Optimization (DPO), discussing how our method situates itself in the context of learning from human preferences.\\

\begin{figure*}[!htb]
\vspace{-20pt}
    \centering
    \includegraphics[scale=0.55]{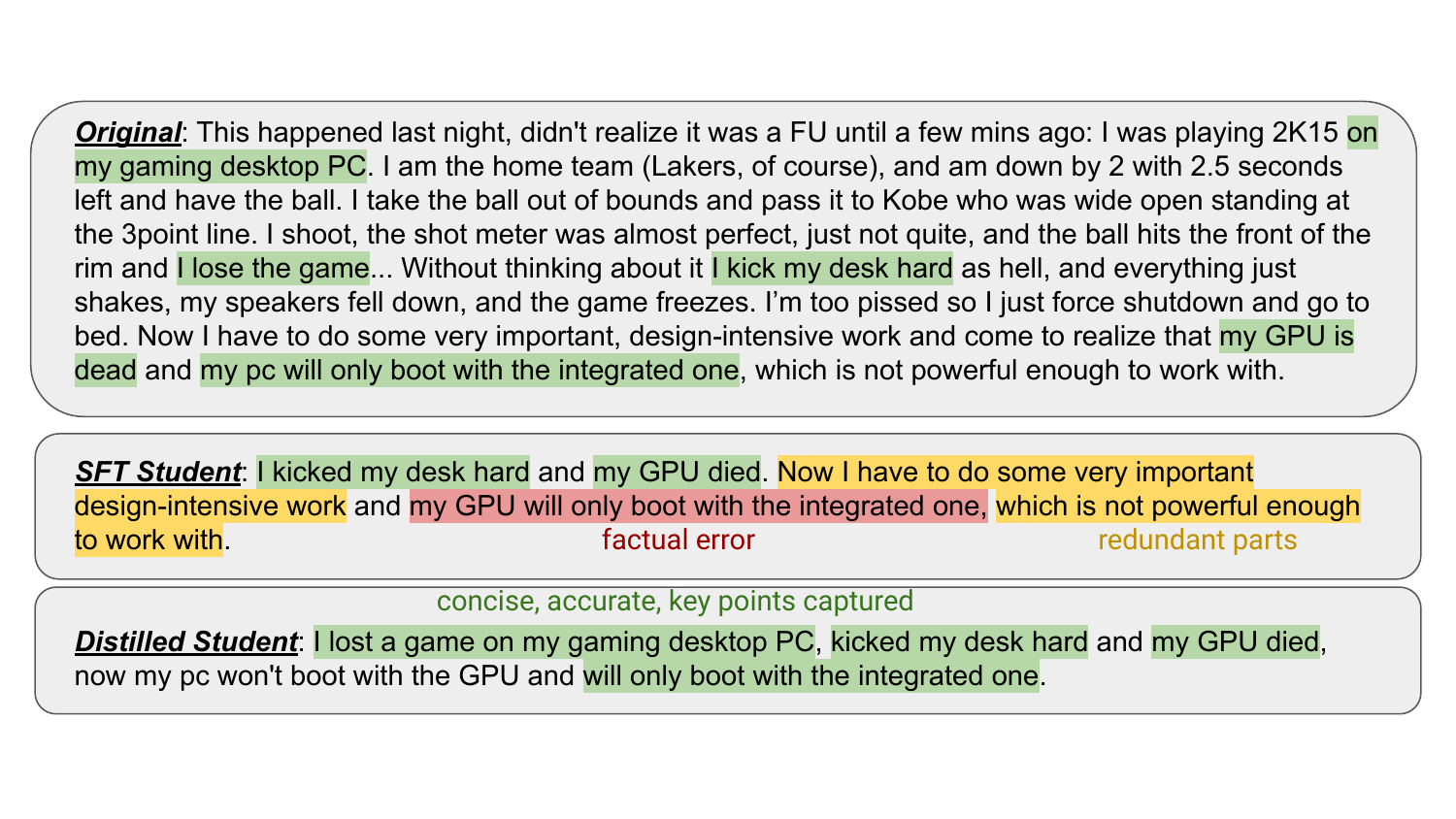}
    \vspace{-30pt}
    \caption{The case study on the TL;DR dataset.}
    \vspace{-8pt}
    \label{fig:case_study}
\end{figure*}

\subsection{Results on More LLMs}
\vspace{-5pt}
\begin{table}[!htb]
\centering
\resizebox{\columnwidth}{!}{
\begin{tabular}{cccc}
\toprule
\textbf{Task} & \textbf{Teacher} & \textbf{Initial} &\textbf{Distill} \\\hline
TL-DR & T5-xxl & 36.67 & 44.46\\
Anthropic-HH & PaLM-2-S & 15.32 & 17.88\\
\bottomrule
\end{tabular}
}
\caption{Win rates of using PALM-2 and T5 models. The teacher model is T5-XXL and PaLM-2-S, respectively, and the student model is T5-large.}
\label{tab:ulm_res}
\end{table}

To demonstrate the generalizability of our framework, we extend our investigations to more LLMs including PaLM 2~\cite{anil2023palm} and T5~\cite{Raffel2019ExploringTL}. 
Specifically,  we employ the T5-XXL as the teacher model for the TL;DR task and the PaLM-2-S as the teacher model for the  Anthropic-HH task. They are distilled into a T5-Large student model in both tasks. The results are presented in Table.~\ref{tab:ulm_res}, and the same conclusion holds that the student distilled with \ours gets a significantly higher win rate compared to the initial student. Our results, spanning LLaMA-2, GPT, T5, and PaLM models, underscore the framework's adaptability across a range of LLM families.

\subsection{Efficacy of General-Purpose Models in Win Rate Calculation}
\begin{table}[!htb]
\centering
\resizebox{\columnwidth}{!}{
\begin{tabular}{ccccc}
\toprule
\textbf{TL;DR} & \textbf{T-S WR} & \textbf{Initial} &\textbf{Baseline{\dag}} &\textbf{\ours} \\\hline
DeBERTa RM & 60.80 & 33.42 & 38.35 & 40.46\\
GPT-3.5-turbo & 63.81 & 31.69 & 37.42 & 38.17\\
\midrule
\textbf{Anthropic-HH} & \textbf{T-S WR} & \textbf{Initial} &\textbf{Baseline*} &\textbf{\ours} \\\hline
DeBERTa RM & 55.23 & 25.63 & 26.24 & 27.74\\
GPT-3.5-turbo & 58.95 & 24.43 & 25.06 & 26.22 \\
\bottomrule
\end{tabular}
}
\vspace{-8pt}
\caption{Comparison of win rates calculated by the task-specific reward model DeBERTa and the general-purpose model GPT-3.5-turbo. The teacher model is LLaMA-2-13B and the student model is LLaMA-2-7B. We present the best baseline methods where \dag~refers to f-distill and * refers to MiniLLM.}
\vspace{-15pt}
\label{tab:autoSxS}
\end{table}
We explore the efficacy of using general-purpose models, specifically GPT-3.5-turbo, for calculating win rates, to provide a parallel evaluation with our deployed task-specific reward models.
Table~\ref{tab:autoSxS} shows that the performance advantage in terms of win rate still holds when evaluated by the general-purpose model. Notably, the T-S WR increases when evaluated by GPT-3.5-turbo, albeit accompanied by a marginal decline in the win rate of model-generated text over the target text. It indicates that compared to the preference made by the reward model, GPT-3.5-turbo favors teacher generation in the teacher-student comparison, while simultaneously showing a slight preference for the target text over model-generated content. Nevertheless, the results affirm that \ours~secures an enhancement in win rate compared to the initial student and the most competitive baseline methods.

% \subsection{Hyperparameter Sensitivity.}
% This study focuses on the sensitivity to the model or training hyperparameters to understand the robustness of our approach across different settings. The studied hyperparameters include the margin $\beta$ in our calibration loss,  the temperature $\tau$ for model generation, and the LoRA rank $r$ in model training. The conclusion is that the LoRA rank $r$ does not impact the final performance much, while the margin $\beta$ and the temperature  $\tau$  slightly impacts the final performance. We detail these experiments in Appendix~\ref{app:hp}.   \\

\subsection{Case Study}
% Through case studies (Figure~\ref{fig:case_study}), we provide qualitative insights into the model's performance, highlighting areas where our approach excels in comparison to baselines.
In Figure~\ref{fig:case_study}, we present a comparative case study from TL;DR to illustrate the efficacy of \ours. We examine the quality of summaries generated by both an SFT (Self-Training) student model and our distilled student model against the original text. Key points from the original narrative are highlighted, providing a benchmark for evaluating the student models' performance. We detail the discussion in Appendix~\ref{app:case}.

\section{Conclusions and Future Work}

In this work, we introduce \ours, a novel preference-based LLM distillation framework that leverages pseudo preference pairs to efficiently transfer knowledge from a large LLM teacher to a compact student model. 
By focusing on the relative ranking of outputs, we allow the student model to learn in a way that is both resource-efficient and aligned with the qualitative nuances of language generation tasks.
Our experiments demonstrate that \ours learns a student model that retains a high level of performance, measured by traditional metrics like ROUGE, and shows significant improvements in terms of generation quality measured by win rate. 
The ablation studies further underscore the importance of our design choices, particularly the use of pseudo-preference pairs over real reward pairs and the implementation of length normalization techniques.

As LLMs continue to expand their role in various applications, methods like \ours that optimize for efficiency and quality without requiring extensive computational resources will become increasingly vital. We hope our work will pave the way for more sustainable and accessible AI language systems in the future.

\section{Limitations}
Potential limitations are: (1) the work depends on the assumption that the teacher model is better than the student model. While the assumption might hold at the beginning stage of student model training, it might not hold when the student model is very carefully trained. This might explicitly create a ceiling for the student model performance. The iterative methods could be considered for future work.
(2) Our approach requires computation resources for conducting bulk inference for both teacher and student models.
(3) To accurately obtain ranking pairs, we can resort to a reward model in the loop. However, it brings computational overhead to the current pipeline.

\bibliography{anthology,custom}
\bibliographystyle{acl_natbib}

\clearpage
\newpage
\appendix

\section{Appendix}

\subsection{Dataset Statistics}\label{app:dataset_stats}
Table~\ref{tab:dataset} shows the statistics of studied datasets.
\begin{table}[!htb]
% \vspace{-10pt}
    \centering
\resizebox{\columnwidth}{!}{
    \begin{tabular}{ccccc}
    \toprule
        Dataset & Train  & Distillation & Dev & Test  \\
        \midrule
        TL;DR  & 56k  & 56k  & 0.5k  & 12k\\
        Anthropic-HH  & 58k & 58k & 6.4k& 6.6k \\
        % ANLI & Multi-class Natural Language Inference & 15,000  & 50,459 &  1,200 & 1,200 \\
        % CIFAR-100 & Image Classification & - & 50,000 & - & 10,000\\
        % Synthetic Dataset & Toy Multi-variate Gaussian&5,000&5,000& 5,000&90,000\\
    \bottomrule
    \end{tabular}
    }
    \caption{Dataset Statistics}
    \label{tab:dataset}
    % \vspace{-10pt}
\end{table}

\begin{table*}
    \centering
    \small
    % \resizebox{\columnwidth}{!}{
    \begin{tabular}{p{5cm}<{\centering}p{6cm}<{\centering}}\toprule
        \textbf{Hyper-parameter} & \textbf{ Search Range}\\\midrule
         Learning Rate & $\{1, 2, 3, 5\} \times10^{-4}$\\
        Batch Size for LLaMA & $\{1, 2, 4, 8, 16\}$\\
         Batch Size for GPT-Neo & $\{4, 8, 16, 32\}$\\
        Temperature $\tau$ &$\{0.1, 0.3, 0.5, 0.7, 0.9\}$\\
        Margin $\beta$  & $\{0.1, 0.2, 0.4, 0.6, 0.8, 1.0, 1.5, 2.0, 2.5\}$\\
        LoRA rank & $\{8, 16, 32\}$\\
    \bottomrule
    \end{tabular}
    % }
    \caption{The search range of hyper-parameters.} \label{tab:hp-search}
\end{table*}

\subsection{Model Details}\label{app:model}
We leverage the open-sourced pre-trained models in our main experiments.
The model cards can be found via \href{https://huggingface.co/meta-llama}{LLaMA-2} and \href{https://huggingface.co/EleutherAI}{GPT-Neo} on Hugging Face.

\subsection{Reward Model}
For the reward model, we use the \href{https://huggingface.co/OpenAssistant/reward-model-deberta-v3-large-v2}{reward-model-deberta-v3-large-v2} released on Hugging Face. It is trained on 
\begin{itemize}
\setlength\itemsep{-0.5em}
    \item  webgpt\_comparisons;
    \item summarize\_from\_feedback;
    \item synthetic-instruct-gptj-pairwise;
    \item anthropic\_hh-rlhf,
\end{itemize}
which covers the related datasets of our studied tasks. Its performance is listed in Table~\ref{tab:app_rm}.
\begin{table}[H]
    \centering
        \resizebox{\columnwidth}{!}{
    \begin{tabular}{ccc}\hline
        Model & Summary & Anthropic RLHF \\\hline
        deberta-v3-large-v2 & 71.47 & 69.25\\\hline
    \end{tabular}
    }
    \caption{Reward Model Performance.}
    \label{tab:app_rm}
\end{table}

\subsection{Case Study}\label{app:case}
 Figure~\ref{fig:case_study} presents the case study on the TL;DR dataset. In the original text, critical details are highlighted in green. Similarly, in the student's summary, critical details are also highlighted in green, with redundant parts in yellow and factual errors in red. Compared to the summaries produced by the SFT student model, those generated by the student model distilled with our method are notably more concise and accurate, capturing a broader range of key points.
 
The original text describes an incident with details. The SFT student's summary, while capturing the essence of the event, includes factual errors and redundant elements. It incorrectly states the GPU's issue, which is a misinterpretation of the original event. Furthermore, it redundantly mentions details not central to the original account's focus. In contrast, the summary produced by our distilled student model is concise and free of factual inaccuracies. It accurately captures the original text and summarizes it into a brief one, maintaining the critical details: the loss of the game, the subsequent damage to the GPU, and the resulting limitation on the computer's functionality. This case study shows an evident improvement after distillation with \ours~, in the model's ability to preserve key information while eliminating superfluous details.

\subsection{Computing Resources}
We test our code on the System Ubuntu 18.04.4
LTS with CPU: Intel(R) Xeon(R) Silver 4214 CPU
@ 2.20GHz and GPU: NVIDIA A100 (80G).
We implement our method using Python 3.9,
PyTorch 2.0.1, and transformers 4.32.

\subsection{Hyper-parameters}\label{app:hp}
We list the search range of hyperparamters in Table~\ref{tab:hp-search}.
The search for batch size and learning rate is applied to all the methods.
And for each baseline, we search for the best baseline-specific hyper-parameters.
For those method-specific hyperparameters, the LoRA rank $r$ does not impact the final performance much, while the margin $\beta$ and the temperature  $\tau$  slightly impacts the final performance, and we choose them carefully. Specifically, we have $\tau=0.7$ and $\beta=1.0$ in the main experiments. 

\subsection{Score Function}
\begin{table*}[!htb]
    \centering
    % \resizebox{\columnwidth}{!}{
\begin{tabular}{lcc|cc}
\hline
 &TL;DR \( \mathcal{L}_{\text{cal}}^{\text{rank}} \) & TL;DR \( \mathcal{L}_{\text{cal}}^{\text{margin}} \) & 
 HH \( \mathcal{L}_{\text{cal}}^{\text{rank}} \)& HH \( \mathcal{L}_{\text{cal}}^{\text{margin}} \)\\
\hline
LLaMA-2 & 40.46 & 40.21 & 27.65 & 27.74 \\
GPT-Neo & 14.70 & 14.76 & 10.38 & 9.94 \\
\hline
\end{tabular}
% }
\caption{Ranking calibration loss v.s. margin calibration loss.}
\end{table*}

The score function in Eq.~\ref{eq:cal_margin} is defined as
\begin{equation}
    s(\hat{y}, y; x) = \sum_{n} F_n \left( e(\hat{y}, y), e(\hat{y}, x) \right),
\end{equation}
where $F_n = 2P_n R_n/(P_n + R_n)$.
The definitions of \( P_n \), \( R_n \), \( F_n \) can be found in (Zhang et al., 2019). When \( n > 1 \), we have
\begin{equation}
    R_n = \frac{1}{|e|} \sum_{i+n \in e} \max_{j+n \in e} e_i^T e_{j+n},
\end{equation}
and
\begin{equation}
    P_n = \frac{1}{|e|} \sum_{j:n \in e} e_i^T e_{j:n+n}.
\end{equation}

This score function measures the similarity between the positive (negative) text and the reference text with negligible computational overhead, because it uses the student model to obtain the representation instead of an external model. By integrating it into Eq.~\ref{eq:cal_margin}, we aim to scale the margin \(\beta\) with the score difference \( s(\hat{y}, y'; x) - s(\hat{y}, y''; x) \). This modulation increases the margin when the positive sequence closely resembles the reference text. A larger margin means that the positive sequences should be preferred more via a higher likelihood in the generation process. This dynamic adjustment of the margin essentially encourages a clearer distinction between positive and negative sequence pairs during training.

For the empirical validation, we choose between the ranking calibration loss and the margin calibration loss based on their performance on the validation set. We report the numbers of each loss in the main experiments setting here.

\end{document}